\documentclass{article}

\usepackage[preprint]{nips_2018}

\usepackage[utf8]{inputenc} % allow utf-8 input
\usepackage[T1]{fontenc}    % use 8-bit T1 fonts
\usepackage{hyperref}       % hyperlinks
\usepackage{url}            % simple URL typesetting
\usepackage{booktabs}       % professional-quality tables
\usepackage{amsfonts}       % blackboard math symbols
\usepackage{nicefrac}       % compact symbols for 1/2, etc.
\usepackage{microtype}      % microtypography
\usepackage{graphicx}
\usepackage{amsmath}
\usepackage{amssymb}

\title{A Unified Framework of Deep Neural Networks by Capsules}

\author{
  Yujian Li\\%\thanks{\emph{https://baike.baidu.com/item/\%E6\%9D\%8E\%E7\%8E\%89\%E9\%89\%B4/1664506?fr=aladdin}.} \\
  College of Computer Science\\
  Faculty of Information Technology\\
  Beijing University of Technology\\
  Beijing, China 100124 \\
  \texttt{liyujian@bjut.edu.cn} \\
  \And
  Chuanhui Shan \\
  College of Computer Science\\
  Faculty of Information Technology\\
  Beijing University of Technology\\
  Beijing, China 100124 \\
  \texttt{chuanhuishan@emails.bjut.edu.cn} \\
}

\begin{document}

\maketitle

\begin{abstract}
With the growth of deep learning, how to describe deep neural networks unifiedly is becoming an important issue. We first formalize neural networks mathematically with their directed graph representations, and prove a generation theorem about the induced networks of connected directed acyclic graphs. Then, we set up a unified framework for deep learning with capsule networks. This capsule framework could simplify the description of existing deep neural networks, and provide a theoretical basis of graphic designing and programming techniques for deep learning models, thus would be of great significance to the advancement of deep learning.
\end{abstract}

\section{Introduction}

Deep learning has made a great deal of success in processing images, audios, and natural languages [1-3], influencing academia and industry dramatically. It is essentially a collection of various methods for effectively training neural networks with deep structures. A neural network is usually regarded as a hierarchical system composed of many nonlinear computing units (or neurons, nodes). The most popular neural network was once multilayer perceptron (MLP) [4]. A MLP consists of an input layer, a number of hidden layers and an output layer, as shown in Figure 1. The depth of it is the number of layers excluding the input layer. If the depth is greater than 2, a neural network is now called “deep”. For training MLPs, backpropagation (BP) is certainly the most well-known algorithm in common use [4], but it seemed to work only for shallow networks. In 1991, Hochreiter indicated that typical deep neural networks (DNNs) suffer from the problem of vanishing or exploding gradients [5]. To overcome training difficulties in DNNs, Hinton et al. started the new field of deep learning in 2006 [6, 7].

Besides deep MLPs, DNNs also include convolutional neural networks (CNNs) and recurrent neural networks (RNNs). Here, we omit RNNs for saving space. Theoretically, a CNN can be regarded as a special MLP or feedforward neural network. It generally consists of an input layer, alternating convolutional and pooling layers, a fully connected layer, and an output layer, as shown in Figure 2. Note that “convolutional layers”  are also called "detection layers", and “pooling layers"  are also called “downsampling layers”. There have been a large number of CNN variants, for example, LeNet [8], AlexNet [1], VGGNet [9], GoogLeNet [10], ResNet [11], Faster R-CNN [12], DenseNet [13], Mask R-CNN [14], YOLO [15], SSD [16], and so on. They not only take the lead in competitions of image classification and recognition as well as object localization and detection [9-12], but also in other applications such as deep Q-networks [17], AlphaGo [18], speech recognition [2], and machine translation [3]. To cope with the disadvantages of CNNs, in 2017 Hinton et al. further proposed a capsule network [19], which is more convincing from the neurobiological point of view. So many deep models are dazzling with different structures. Some of them have added shortcut connections, parallel connections, and even nested structures to traditional layered structures. How to establish a unified framework for DNNs is becoming a progressively important issue in theory. We are motivated to address it.

\begin{figure}
  \centering
  % \fbox{\rule[-.5cm]{0cm}{4cm} \rule[-.5cm]{4cm}{0cm}}
  \includegraphics[width=3in,height=1.3in]{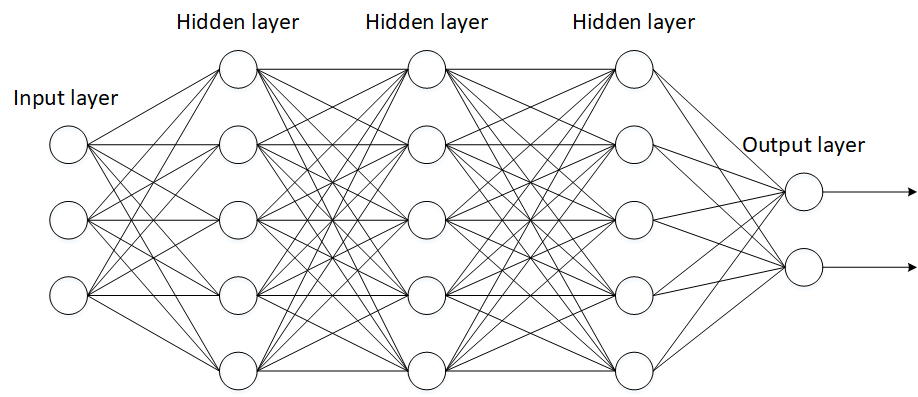}
  \caption{The structure of a MLP.}
\end{figure}

\begin{figure}
  \centering
  % \fbox{\rule[-.5cm]{0cm}{4cm} \rule[-.5cm]{4cm}{0cm}}
  \includegraphics[width=4.5in,height=1.2in]{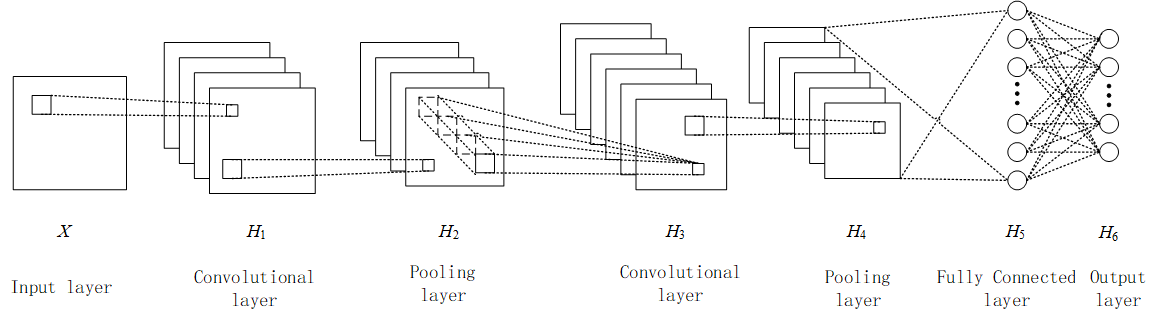}
  \caption{The structure of a CNN.}
\end{figure}

This paper is organized as follows. In Section 2, we propose a mathematical definition to formalize neural networks, give their directed graph representations, and prove a generation theorem about the induced networks of connected directed acyclic graphs. In Section 3, we use the concept of capsule to extend neural networks, define an induced model for capsule networks, and establish a unified framework for deep learning with a universal backpropagation algorithm. Finally, in Section 4 we make a few conclusions to summarize the significance of the capsule framework to advance deep learning in theory and application.

\section{Formalization of Neural networks}

\subsection{Mathematical definition}

A neural network is a computational model composed of nodes and connections. Nodes are divided into input nodes and neuron nodes. Input nodes can be represented by real variables, e.g. $x_{1},x_{2},\cdots,x_{n}$ . The set of input nodes is denoted as $X=\{x_{1},x_{2},\cdots,x_{n}\}$ . A neuron node can receive signals through connections both from input nodes and the outputs of other neuron nodes, and perform a weighted sum of these signals for a nonlinear transformation. Note that the weight measures the strength of a connection, and the nonlinear transformation is the effect of an activation function. Let $F$ be a set of activation functions, such as sigmoid, tanh, ReLU, and so on.

On $X$ and $F$, a neural network can be formally defined as a 4-tuple $net=(S,H,W,Y)$, where $S$ is a set of input nodes, $H$ is a set of neuron nodes, $W$ is a set of weighting connections, and $Y$ is a set of outputs. The neural network is recursively generated by four basic rules as follows:

1) \textbf{Rule of variable}. For any $z\in X$, let $y_z = z$. If $S=\{z\}$, $H=\emptyset$, $W=\emptyset$, $Y=\{y_z\}$, then the 4-tuple $net=(S,H,W,Y)$ is a neural network.

2) \textbf{Rule of neuron}. For any nonempty subset $S\subseteq X$, $\forall f \in F$, $\forall b \in \mathbb{R}$, construct a node $h\not\in X$ that depends on $(f,b)$ and select a set of weighting connections $w_{x_i\rightarrow h} (x_i\in S)$. Let $y_h = f(\sum_{x_i \in S} {w_{x_i\rightarrow h}x_i}+b)$ be the output of node $h$. If $H=\{h\}$, $W=\{w_{x_i\rightarrow h} |x_i\in S\}$, and $Y=\{y_h\}$, then $net=(S,H,W,Y)$ is a neural network.

3) \textbf{Rule of growth}. Suppose $net=(S,H,W,Y)$ is a neural network. For any nonempty subset $N\subseteq S\cup H$, $\forall f\in F$, $\forall b\in \mathbb{R}$, construct a node $h\not\in S\cup H$ that depends on $(f,b)$ and select a set of weighting connections  $w_{z_j\rightarrow h}(z_j\in N)$. Let $y_h = f(\sum_{z_j\in N} {w_{z_j\rightarrow h}y_{z_j}}+b)$ be the output of node $h$. If $S'=S$, $H'=H\cup \{h\}$, $W'=W\cup \{w_{z_j\rightarrow h}|z_j \in N\}$, and $Y'=Y\cup \{y_h\}$, then $net'=(S',H',W',Y')$ is also a neural network.

4) \textbf{Rule of convergence}. Suppose $net_k=(S_k,H_k,W_k,Y_k)(1\leq k \leq K)$ are $K$ neural networks, satisfying that $\forall 1\leq i\neq j \leq K$, $(S_i \cup H_i) \cap (S_j\cup H_j)=\emptyset$. For any nonempty subsets $A_k \subseteq S_k \cup H_k(1\leq k \leq K)$, $N=\bigcup_{k=1}^K {A_k}$, $\forall f\in F$, $\forall b \in \mathbb{R}$, construct a node $h\not\in \bigcup_{k=1}^K (S_k\cup H_k)$ that depends on $(f,b)$, select a set of weighting connections $w_{z\rightarrow h}(z\in N)$. Let $y_h=f(\sum_{z\in N} {w_{z\rightarrow h}y_z}+b)$ be the output of the node $h$. If $S=\bigcup_{k=1}^K S_k$, $H=(\bigcup_{k=1}^K H_k)\cup \{h\}$, $W=(\bigcup_{k=1}^K W_k)\cup \{w_{z\rightarrow h}|z\in N\}$, and $Y=(\bigcup_{k=1}^K Y_k)\cup \{y_h\}$, then $net=(S,H,W,Y)$ is also a neural network.

Among the four generation rules, it should be noted that the rule of neuron is not independent. This rule can be derived from the rule of variable and the rule of convergence. Moreover, the weighting connection $w_{z\rightarrow h}$ should be taken as a combination of the weight and the connection, rather than just the weight. Additionally, if a node $h$ depends on $(f,b)$, $f$ is called the activation function of $h$, and $b$ is called the bias of $h$.

\subsection{Directed graph representation}

Let $X$ be a set of real variables and $F$ be a set of activation functions. For any neural network $net=(S,H,W,Y)$ on $X$ and $F$, a directed acyclic graph $G_{net}=(V,E)$ can be constructed with the vertex set $V=S\cup H$ and the directed edge set $E=\{z\rightarrow h|w_{z\rightarrow h} \in W\}$. $G_{net}=(V,E)$ is called the directed graph representation of $net=(S,H,W,Y)$. Two cases of the representation generation are discussed in the following.

\textbf{1)The case of $X=\{x_1\}$}

Using the rule of variable, for $x_1 \in X$, let $y_{x_1}=x_1$. If $S=\{x_1\}$, $H=\emptyset$, $W=\emptyset$, and $Y=\{y_{x_1}\}$, then $net=(S,H,W,Y)$ is a neural network. Since this network has only one input node without any function for nonlinear transformation, it is also called a trivial network, as shown in Figure 3(a). Using the rule of neuron, for a nonempty subset $S=\{x_1\} \subseteq X$, $\forall f\in F$, $\forall b \in \mathbb{R}$, construct a node $h_1 \not\in S$ that depends on $(f,b)$, select a weighting connection $w_{x_1\rightarrow h_1}$, and let $y_{h_1}=f(w_{x_1\rightarrow h_1}x_1 +b)$. If $H=\{h_1\}$, $W=\{w_{x_1\rightarrow h_1}\}$, and $Y=\{y_{h_1}\}$, then $net=(S,H,W,Y)$ is a neural network, which has one input and one neuron. It is also called a 1-input-1-neuron network, as shown in Figure 3(b). Using the rule of growth on the network, three new neural networks with different structures can be generated, as shown in Figures 4(a-c). Likewise, they are called 1-input-2-neuron networks. Using the rule of growth on the three networks, twenty-one new neural networks with different structures can be totally generated further. Seven out of them for Figure 4(a) are displayed in Figures 5(a-g). They are called 1-input-3-neuron networks.

\begin{figure}
  \centering
  % \fbox{\rule[-.5cm]{0cm}{4cm} \rule[-.5cm]{4cm}{0cm}}
  \includegraphics[width=2.3in,height=0.5in]{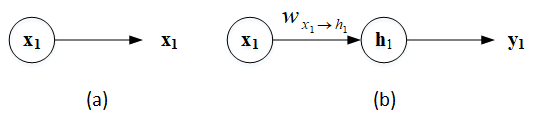}
  \caption{(a)A trivial network; (b)A 1-input-1-neuron network.}
\end{figure}

\begin{figure}
  \centering
  % \fbox{\rule[-.5cm]{0cm}{4cm} \rule[-.5cm]{4cm}{0cm}}
  \includegraphics[width=4.7in,height=0.9in]{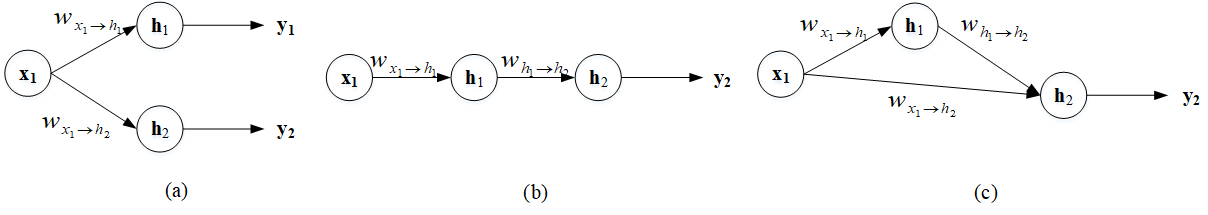}
  \caption{Three 1-input-2-neuron networks.}
\end{figure}

\begin{figure}
  \centering
  % \fbox{\rule[-.5cm]{0cm}{4cm} \rule[-.5cm]{4cm}{0cm}}
  \includegraphics[width=5.5in,height=2in]{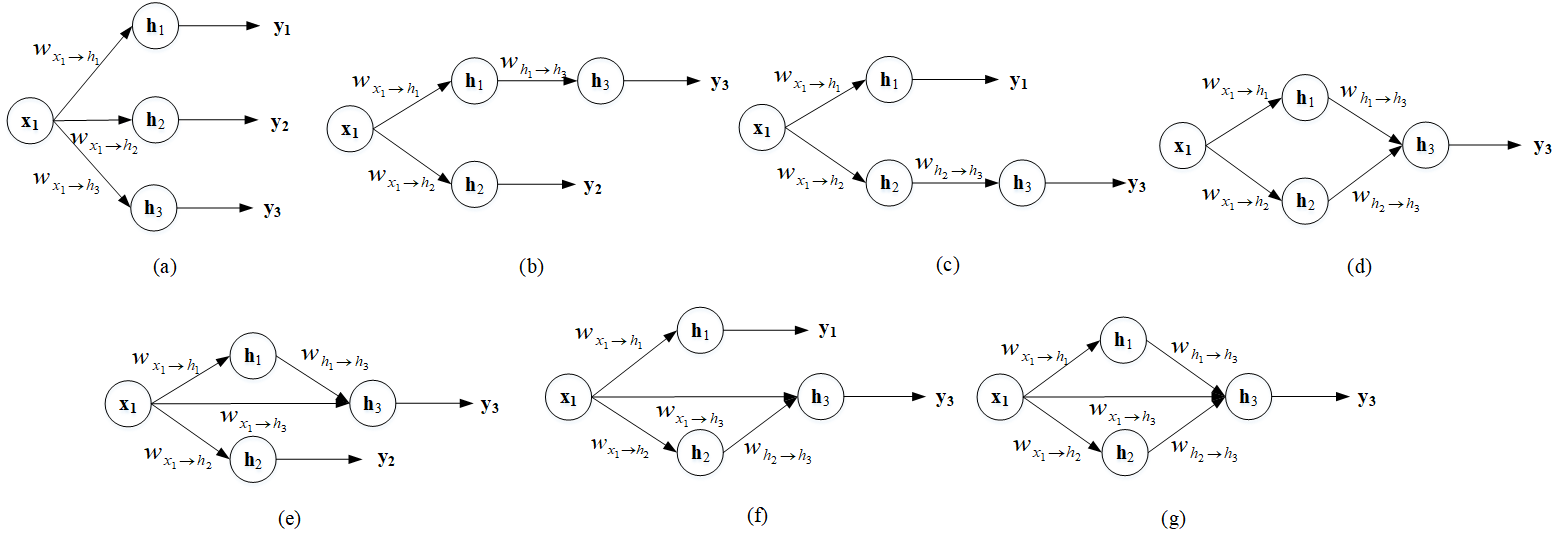}
  \caption{Seven 1-input-3-neuron networks.}
\end{figure}

\textbf{2)The case of $X=\{x_1,x_2\}$}

Using the rule of variable, for $x_1, x_2\in X$, let $y_{x_1}=x_1$ and $y_{x_2}=x_2$. If $S_1=\{x_1\}$, $S_2=\{x_2\}$, $H_1=H_2=\emptyset$, $W_1=W_2=\emptyset$, $Y_1=\{y_{x_1}\}$, and $Y_2=\{y_{x_2}\}$, then $net_1=(\{x_1\},\emptyset,\emptyset,\{y_{x_1}\})$ and $net_2=(\{x_2\},\emptyset,\emptyset,\{y_{x_2}\})$ are neural networks. Obviously, both of them are trivial networks. Using the rule of neuron, for a nonempty subset $S\subseteq X$, if $S=\{x_1\}$ or $S=\{x_2\}$, the neural network can be similarly constructed with the case of $X=\{x_1\}$.

If $X=\{x_1,x_2\}$, $\forall f\in F$, $\forall b \in \mathbb{R}$, construct a node $h_1\not\in S$ that depends on $(f,b)$, select a set of weighting connections $w_{x_i\rightarrow h_1}(x_i \in S)$ and let $y_{h_1}=f(\sum_{x_i\in S}{w_{x_i\rightarrow h_1}x_i}+b)$. If $H=\{h_1\}$, $W=\{w_{x_1\rightarrow h_1}, w_{x_2\rightarrow h_1}\}$, and $Y=\{y_{h_1}\}$, then $net=(S,H,W,Y)$ is a neural network. This is a 2-input-1-neuron network, as depicted in Figure 6. Using the rule of growth on this network, seven 2-input-2-neuron networks with different structures can be generated, as shown in Figures 7(a-g).
\begin{figure}
  \centering
  % \fbox{\rule[-.5cm]{0cm}{4cm} \rule[-.5cm]{4cm}{0cm}}
  \includegraphics[width=1.2in,height=0.6in]{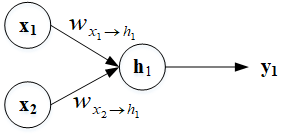}
  \caption{A 2-input-1-neuron network.}
\end{figure}
\begin{figure}
  \centering
  % \fbox{\rule[-.5cm]{0cm}{4cm} \rule[-.5cm]{4cm}{0cm}}
  \includegraphics[width=5.5in,height=1.8in]{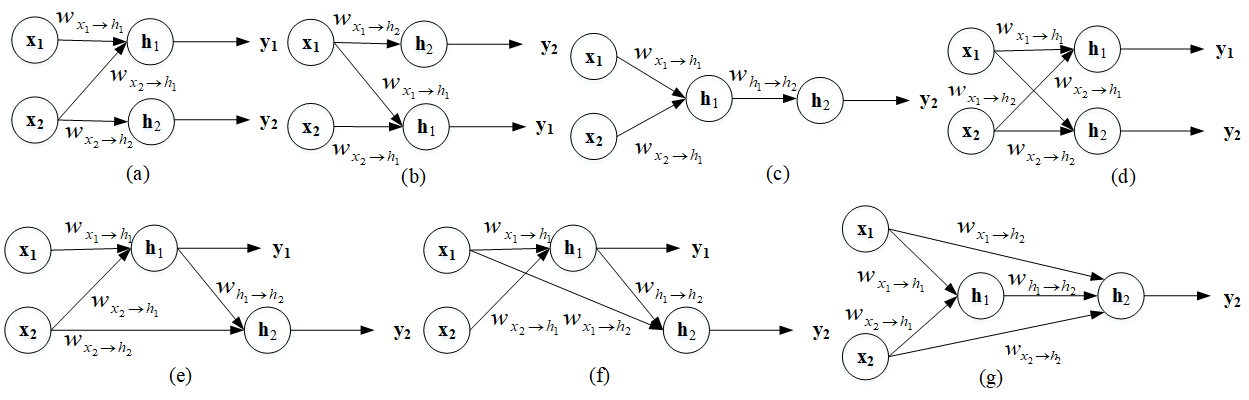}
  \caption{Seven 2-input-2-neuron networks.}
\end{figure}

Finally, the rule of convergence is necessary. In fact, it cannot generate all neural networks only using the three rules of variable, neuron and growth. For example, the network in Figure 8(c) cannot be generated without using the rule of convergence on the two in Figures 8(a-b).
\begin{figure}
  \centering
  % \fbox{\rule[-.5cm]{0cm}{4cm} \rule[-.5cm]{4cm}{0cm}}
  \includegraphics[width=4.3in,height=0.9in]{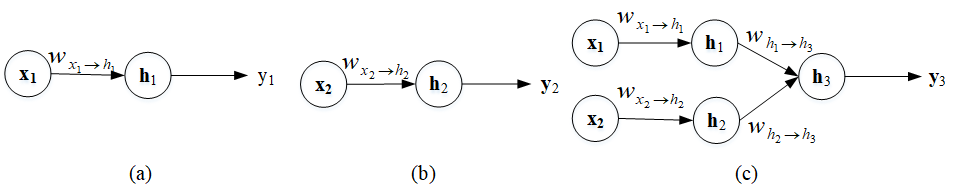}
  \caption{A necessary explanation for the rule of convergence.}
\end{figure}

\subsection{Induced network and its generation theorem}

Suppose $G=(V,E)$ is a connected directed acyclic graph, where $V$ denotes the vertex set and $E$ denotes the directed edge set. For any vertex $h\in V$, let $IN_h=\{z|z\in V, z\rightarrow h \in E\}$ be the set of vertices each with a directed edge to $h$, and $OUT_h=\{z|z\in V,h\rightarrow z \in E\}$ be the set of vertices for $h$ to have directed edges each to. If $IN_h=\emptyset$, then $h$ is called an input node of $G$. If $OUT_h=\emptyset$, then $h$ is called an output node of $G$.  Otherwise, $h$ is called a hidden node of $G$. Let $X$ stand for the set of all input nodes, $O$ for the set of all output nodes, and $M$ for the set of all hidden nodes. Obviously, $V=X\cup M\cup O$, and $M=V-X\cup O$.

Furthermore, let $y_h$ be the output of node $h$, and $w_{z\rightarrow h}$ be the weighting connection from $z$ to $h$. Then, a computational model of graph $G$ can be defined as follows:

\hspace{0.4cm}1) $\forall z \in X$, $y_z=z$.

\hspace{0.4cm}2) $\forall h\in M\cup O$, select $f\in F$ and $b \in \mathbb{R}$ to compute $y_h=f(\sum_{z\in IN_h}{w_{z\rightarrow h}y_z}+b)$.

If $S=X$, $H=M\cup O$, $W=\{w_{z\rightarrow h}|z\rightarrow h \in E\}$, and $Y=\{y_h|h\in V\}$, then $net_G = (S,H,W,Y)$ is called an induced network of graph $G$. The following generation theorem holds on the induced network.

\textbf{Generation Theorem:} For any connected directed acyclic graph $G=(V,E)$, its induced network $net_G$ is a neural network that can be recursively generated by the rules of variable, neuron, growth, and convergence.

\textbf{Proof:} By induction on $|V|$ (i.e. number of vertices), we prove the theorem as follows.\\
1) When $|V|=1$, we have $|X|=1$ and $|O|=0$, so the induced network $net_G$ is a neural network that can be generated directly by the rule of variable.\\
2) When $|V|=2$, we have $|X|=1$ and $|O|=1$, so the induced network $net_G$ is a neural network that can be generated directly by the rule of growth.\\
3) Assume that the theorem holds for $|V|\leq n$. When $|V|= n+1\geq 3$, the induced network $net_G$ has at least one output node $h\in O$. Let $E_h=\{z\rightarrow h\in E\}$ denote the set of edges heading to the node $h$. Moreover, let $V'=V-\{h\}$ and $E'=E-E_h$. Based on the connectedness of $G'=(V',E')$, we have two cases to discuss in the following:
\begin{enumerate}
\item[i)] If $G'=(V',E')$ is connected, then applying the induction assumption for $|V'|\leq n$, the induced network $net_{G'}=(S',H',W',Y')$ can be recursively generated by the rules of variable, neuron, growth, and convergence. Let $N=IN_h$. In $net_G=(S,H,W,Y)$, we use $f\in F$ and $b\in \mathbb{R}$ to stand for the activation function and bias of node $h$, and $w_{z\rightarrow h}(z\in N)$ for the weighting connection from node $z$ to the node $h$. Then, $net_G$ can be obtained by using the rule of growth on $net_{G'}$, to generate the node $h$ and its output $y_h = f(\sum_{z\in N}{w_{z\rightarrow h}y_z}+b)$.
\item[ii)] Otherwise, $G'$ comprises a number of disjoint connected components $G_k=(V_k,E_k)(1\leq k \leq K)$. Using the induction assumption for $|V_k|\leq n(1\leq k \leq K)$, the induced network $net_{G_k}=(S_k,H_k,W_k,Y_k)$ can be recursively generated by the rules of variable, neuron, growth, and convergence. Let $A_k=(S_k\cup H_k)\cap IN_h$, and $N= \bigcup_{k=1}^K{A_k}$. In $net_G=(S,H,W,Y)$, we use $f\in F$ and $b\in \mathbb{R}$ to stand for the activation function and bias of the node $h$, and $w_{z\rightarrow h}(z\in N)$ for the weighting connection from node $z$ to node $h$. Then, $net_G$ can be obtained by using the rule of convergence on $net_{G_k}(1\leq k \leq K)$, to generate the node $h$ and its output $y_h = f(\sum_{z\in N}{w_{z\rightarrow h}y_z}+b)$.
\end{enumerate}
As a result, the theorem always holds.

\section{Capsule framework of Deep learning}

\subsection{Mathematical definition of capsules}

In 2017, Hinton et al. pioneered the idea of capsules and considered a nonlinear “squashing” capsule [19]. From the viewpoint of mathematical models, a capsule is essentially an extension of the traditional activation function. It is primarily defined as an activation function with a vector input and a vector output. More generally, a capsule can be an activation function with a tensor input and a tensor output.

\begin{figure}
  \centering
  % \fbox{\rule[-.5cm]{0cm}{4cm} \rule[-.5cm]{4cm}{0cm}}
  \includegraphics[width=3in,height=0.9in]{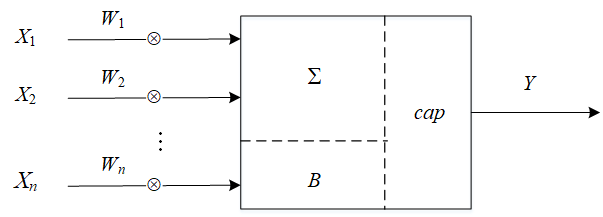}
  \caption{Mathematical model of a general capsule.}
\end{figure}

As shown in Figure 9, a general capsule may have $n$ input tensors $X_1,X_2,\cdots,X_n$, $n$ weight tensors $W_1,W_2,\cdots,W_n$, and a capsule bias $B$, and $n$ weighting operations $\otimes_1,\otimes_2,\cdots,\otimes_n$. Note that a weighting operation may be taken as an identity transfer, a scalar multiplication, a vector dot product, a matrix multiplication, a convolution operation, and so on. Meantime, $W_i\otimes_iX_i(1\leq i \leq n)$ and $B$ must be tensors with the same dimension. The total input of the capsule is $U=\sum_i{W_i\otimes_iX_i}+B$, and the output $Y$ is a tensor computed by a nonlinear capsule function $cap$, namely,

\begin{equation}
  Y=cap(U)=cap(\sum_i{W_i\otimes_iX_i}+B).
\end{equation}

For convenience, we use $\mathcal{F}$ to stand for a nonempty set of capsule functions, and $\mathbb{T}$ for the set of all tensors.

\subsection{Capsule Networks}

Suppose $\mathcal{G}=(\mathcal{V},\mathcal{E})$ is a connected directed acyclic graph, where $\mathcal{V}$ denotes the vertex set and $\mathcal{E}$ denotes the directed edge set. For any vertex $H\in \mathcal{V}$, let $IN_H$ be the set of vertices each with a directed edge to $H$, and $OUT_H$ be the set of vertices for $H$ to have a directed edge each to. If $IN_H=\emptyset$, then $H$ is called an input node of $\mathcal{G}$. If $OUT_H=\emptyset$, then $H$ is called an output node of $\mathcal{G}$. Otherwise, $H$ is called a hidden node of $\mathcal{G}$. Let $\mathcal{X}$ stand for the set of all input nodes, $\mathcal{O}$ for the set of all output nodes, and $\mathcal{M}$ for the set of all hidden nodes. Obviously, $\mathcal{V}=\mathcal{X}\cup \mathcal{M}\cup \mathcal{O}$, and $\mathcal{M}=\mathcal{V}-\mathcal{X}\cup \mathcal{O}$.

Furthermore, let $Y_H$ be the output of node $H$, and $(W_{Z\rightarrow H},\otimes_{Z\rightarrow H})$ be the tensor-weighting connection from $Z$ to $H$. If $\forall H\in \mathcal{M}\cup \mathcal{O}, \forall Z \in IN_H$, $W_{Z\rightarrow H}\otimes_{Z\rightarrow H}Y_Z$ and $B$ are tensors with the same dimension, then a tensor-computational model of graph $\mathcal{G}$ can be defined as follows:

\hspace{0.4cm}1) $\forall Z \in \mathcal{X}$, $Y_Z=Z$.

\hspace{0.4cm}2) $\forall H\in \mathcal{M}\cup \mathcal{O}$, select $cap\in \mathcal{F}$ and $B \in \mathbb{T}$ to compute $Y_H=cap(\sum_{Z\in IN_H}{W_{Z\rightarrow H}\otimes_{Z\rightarrow H}Y_Z}+B)$.

If $\mathcal{S}=\mathcal{X}$, $\mathcal{H}=\mathcal{M}\cup \mathcal{O}$, $\mathcal{W}=\{(W_{Z\rightarrow H},\otimes_{Z\rightarrow H})|Z\rightarrow H\in \mathcal{E}\}$, and $\mathcal{Y}=\{Y_H|H\in \mathcal{V}\}$, then $net_{\mathcal{G}}=(\mathcal{S},\mathcal{H},\mathcal{W},\mathcal{Y})$ is called a tensor-induced network of graph $\mathcal{G}$. This network is also called a capsule network.

\begin{figure}
  \centering
  % \fbox{\rule[-.5cm]{0cm}{4cm} \rule[-.5cm]{4cm}{0cm}}
  \includegraphics[width=3.5in,height=0.4in]{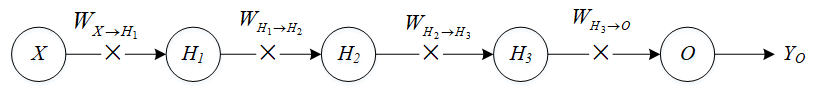}
  \caption{The capsule structure of a MLP.}
\end{figure}

Using a capsule network, a MLP can be simplified as a directed acyclic path of capsules. For example, the MLP in Figure 1 has five layers: an input layer, three hidden layers, and an output layer. On the whole, each layer could be thought of as a capsule. Let $X=(x_1,x_2,\cdots,x_5)^T$ stand for the input capsule node, $H_i=(cap_i,B_i)(i=1,2,3)$ for the hidden capsule nodes, and $O=(cap_4,B_4)$ for the output capsule node. Note that capsule function $cap_i$ and capsule bias $B_i$ are defined by the elementwise activation function and the bias vector respectively of the corresponding layer in the MLP. If the weighting operations $\otimes_{X\rightarrow H_1}$, $\otimes_{H_1\rightarrow H_2}$, $\otimes_{H_2\rightarrow H_3}$, and $\otimes_{H_3\rightarrow O}$ are all taken as matrix multiplication “$\times$”, then we have $(W_{X\rightarrow H_1},\otimes_{X\rightarrow H_1})=((w_{m,n}^{X\rightarrow H_1})_{7\times5},\times)$, $(W_{H_1\rightarrow H_2},\otimes_{H_1\rightarrow H_2})=((w_{m,n}^{H_1\rightarrow H_2})_{7\times7},\times)$, $(W_{H_2\rightarrow H_3},\otimes_{H_2\rightarrow H_3})=((w_{m,n}^{H_2\rightarrow H_3})_{7\times7},\times)$ and $(W_{H_3\rightarrow O},\otimes_{H_3\rightarrow O})=((w_{m,n}^{H_3\rightarrow O})_{4\times7},\times)$, which are the tensor-weighting connections from $X$ to $H_1$, $H_1$ to $H_2$, $H_2$ to $H_3$ and $H_3$ to $O$. Finally, let $Y_{H_i}(i=1,2,3)$ stand for the output vector of $H_i$, and $Y_O$ for the output vector of $O$. Setting $Y_{H_O}=X$ and $Y_{H_4}=Y_O$, we obtain $Y_{H_i}=cap_i(W_{H_{i-1}\rightarrow H_i}\times Y_{H_{i-1}}+B_i)$. Therefore, the capsule structure of the MLP is a directed acyclic path, as displayed in Figure 10.

Besides MLPs, capsule networks can also be used to simplify the structures of other DNNs. Let us consider the CNN in Figure 2. This CNN has 7 layers: one input layer, two convolutional layers, two downsampling (pooling) layers, one fully connected layer, and one output layer. On the whole, each of the layers could be thought of as a capsule. Let $X$ stand for the input capsule node, $H_i=(cap_i,B_i)(i=1,\cdots,5)$ for the hidden capsule nodes, and $O=(cap_6,B_6)$ for the output capsule node. Note that $cap_1$ and $cap_3$ are capsule functions defined by elementwise ReLUs. $cap_2$ and $cap_4$ are capsule functions defined by downsampling “$\downarrow$”. $cap_5$ is an identity function. $cap_6$ is a capsule function defined by softmax. In addition, $B_i(i=1,\cdots,6)$ are capsule biases each defined by the bias tensor of the corresponding layer in the CNN. Let both $\otimes_{X\rightarrow H_1}$ and $\otimes_{H_2\rightarrow H_3}$ be the convolution operation “$\ast$”, both $\otimes_{H_1\rightarrow H_2}$ and $\otimes_{H_3\rightarrow H_4}$ be the  identity transfer “$\rightarrow $”, $\otimes_{H_4\rightarrow H_5}$ be the tensor-reshaping operation “$\triangleleft$”, and $\otimes_{H_5\rightarrow O}$ be the matrix multiplication “$\times$”. Then, $(W_{X\rightarrow H_1},\otimes_{X\rightarrow H_1})=(W_{X\rightarrow H_1},\ast)$, $(W_{H_1\rightarrow H_2},\otimes_{H_1\rightarrow H_2})=("",\rightarrow)$, $(W_{H_2\rightarrow H_3},\otimes_{H_2\rightarrow H_3})=(W_{H_2\rightarrow H_3},\ast)$, $(W_{H_3\rightarrow H_4},\otimes_{H_3\rightarrow H_4})=("",\rightarrow)$, $(W_{H_4\rightarrow H_5},\otimes_{H_4\rightarrow H_5})=("",\triangleleft)$, and $(W_{H_5\rightarrow O},\otimes_{H_5\rightarrow O})=(W_{H_5\rightarrow O},\times)$, which are the tensor-weighting connections from $X$ to $H_1$, $H_1$ to $H_2$, $H_2$ to $H_3$, $H_3$ to $H_4$, $H_4$ to $H_5$, and $H_5$ to $O$. Finally, let $Y_{H_i}(i=1,2,3,4,5)$ stand for the output tensor of $H_i$, and $Y_O$ for the output tensor of $O$. This leads to the following computations:

\begin{equation}
\begin{cases}
  Y_{H_1} = cap_1(W_{X\rightarrow H_1}\ast X+B_1) = \textrm{ReLU}(W_{X\rightarrow H_1}\ast X+B_1), \\
  Y_{H_2} = cap_2(W_{H_1\rightarrow H_2}\otimes_{H_1\rightarrow H_2} Y_{H_1}+B_2) = cap_2(\rightarrow Y_{H_1}+B_2) = \downarrow Y_{H_1}+B_2, \\
  Y_{H_3} = cap_3(W_{H_2\rightarrow H_3}\ast X+B_3) = \textrm{ReLU}(W_{H_2\rightarrow H_3}\ast Y_{H_2}+B_3), \\
  Y_{H_4} = cap_4(W_{H_3\rightarrow H_4}\otimes_{H_3\rightarrow H_4} Y_{H_3}+B_4) = cap_4(\rightarrow Y_{H_3}+B_4) = \downarrow Y_{H_3}+B_4, \\
  Y_{H_5} = cap_5(\triangleleft Y_{H_4}+B_5) = \triangleleft Y_{H_4}, \\
  Y_O = cap_6(W_{H_5\rightarrow O}\times Y_{H_5}+B_6) = softmax(W_{H_5\rightarrow O}\times Y_{H_5}+B_6).
\end{cases}
\end{equation}

Therefore, the capsule structure of the CNN is also a directed acyclic path, as depicted in Figure 11.

\begin{figure}
  \centering
  % \fbox{\rule[-.5cm]{0cm}{4cm} \rule[-.5cm]{4cm}{0cm}}
  \includegraphics[width=4.5in,height=0.4in]{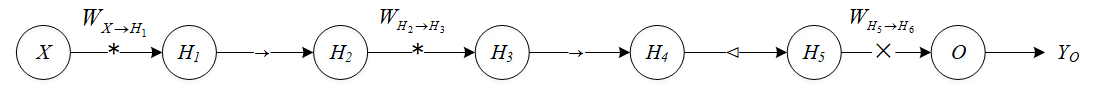}
  \caption{The Capsule structure of a CNN, with “$\ast$” standing for convolution, “$\rightarrow$” for identity transfer, “$\triangleleft$” for tensor reshaping, and “$\times$” for matrix multiplication.}
\end{figure}

Besides simplifying the description of existing DNNs, the capsule networks can also be used to graphically design a variety of new structures for complex DNNs, such as displayed in Figure 12.

\begin{figure}
  \centering
  % \fbox{\rule[-.5cm]{0cm}{4cm} \rule[-.5cm]{4cm}{0cm}}
  \includegraphics[width=3.5in,height=1.8in]{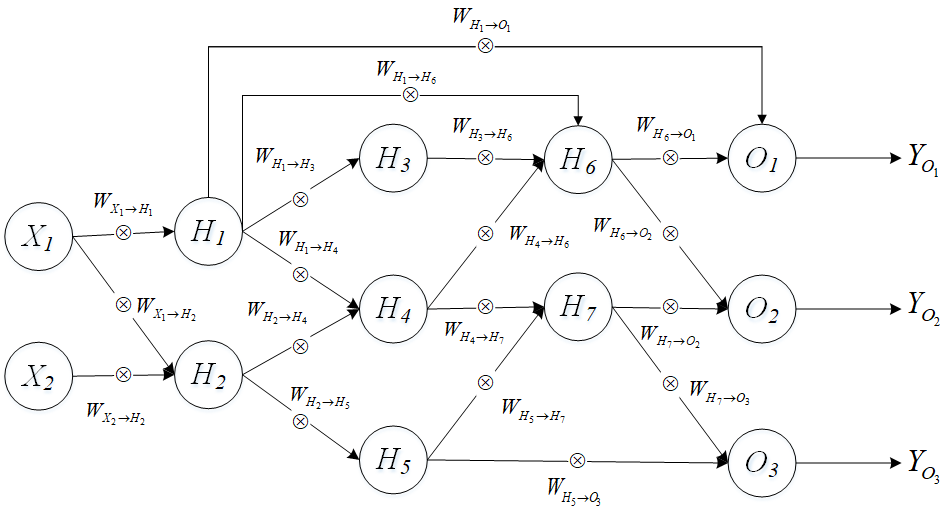}
  \caption{Structure of a general capsule network.}
\end{figure}

\subsection{Universal backpropagation of capsule networks}

Suppose $\mathcal{G}=(\mathcal{V},\mathcal{E})$ is a connected directed acyclic graph. Let $\mathcal{X}=\{X_1,X_2,\cdots,X_n\}$ stand for the set of all input nodes, $\mathcal{O}=\{O_1,O_2,\cdots,O_m\}$ for the set of all output nodes, and $\mathcal{M}=\mathcal{V}-\mathcal{X}\cup \mathcal{O}=\{H_1,H_2,\cdots,H_l\}$ for the set of all hidden nodes. $net_{\mathcal{G}}=(\mathcal{S},\mathcal{H},\mathcal{W},\mathcal{Y})$ is an tensor-induced network of graph $\mathcal{G}$.  This is also a capsule network. If the number of nodes $|\mathcal{S}\cup \mathcal{H}|\geq 2$, then for $\forall H\in \mathcal{H}$,

\begin{equation}
\begin{cases}
  U_H = \sum_{Z\in IN_H}{W_{Z\rightarrow H}\otimes_{Z\rightarrow H}Y_Z+B_H}, \\
  Y_H = cap_H(U_H)=cap_H(\sum_{Z\in IN_H}{W_{Z\rightarrow H}\otimes_{Z\rightarrow H}Y_Z+B_H}).
\end{cases}
\end{equation}

For any output node $H\in \mathcal{O}$, let $Y_H$ and $T_H$ be its actual output and expected output for input $\mathcal{X}$, respectively. The loss function between them is defined as $L_H=Loss(Y_H,T_H)$. Accordingly, we have the total loss function $L=\sum_{H\in \mathcal{O}}L_H$. Let $\delta_H=\frac{\partial L}{\partial U_H}$ denote the backpropagated error signal (or sensitivity) for capsule node $H$. By the chain rule, we further obtain:

\begin{equation}
\forall H\in \mathcal{O},
\begin{cases}
  \delta_H & = \frac{\partial L}{\partial U_H} = \frac{\partial Loss(Y_H,T_H)}{\partial Y_H}\cdot \frac{\partial cap_H}{\partial U_H}, \\
  \frac{\partial L}{\partial B_H} & = \frac{\partial L}{\partial U_H}\cdot \frac{\partial U_H}{\partial B_H} = \delta_H, \\
  \frac{\partial L}{\partial W_{Z\rightarrow H}} & = \frac{\partial L}{\partial U_H}\cdot \frac{\partial U_H}{\partial W_{Z\rightarrow H}} =\delta_H \cdot \frac{\partial U_H}{\partial W_{Z\rightarrow H}}.
\end{cases}
\end{equation}

\begin{equation}
\forall H\in \mathcal{M},
\begin{cases}
  \delta_H & = \frac{\partial L}{\partial U_H} = \sum_{P\in OUT_H}{\frac{\partial L}{\partial U_P}\cdot \frac{\partial U_P}{\partial Y_H}\cdot \frac{\partial Y_H}{\partial U_H}} \\
& = \sum_{P\in OUT_H}{\delta_P \cdot \frac{\partial U_P}{\partial Y_H}\cdot \frac{\partial cap_H}{\partial U_H}}, \\
  \frac{\partial L}{\partial B_H} & = \frac{\partial L}{\partial U_H}\cdot \frac{\partial U_H}{\partial B_H}=\delta_H, \\
  \frac{\partial L}{\partial W_{Z\rightarrow H}} & = \frac{\partial L}{\partial U_H}\cdot \frac{\partial U_H}{\partial W_{Z\rightarrow H}}=\delta_H \cdot \frac{\partial U_H}{\partial W_{Z\rightarrow H}}.
\end{cases}
\end{equation}

Note that in formulae (4)-(5), $\frac{\partial cap_H}{\partial U_H}$ depends on the specific form of capsule function $cap_H$. For example, when $cap_H$ is an elementwise sigmoid function, the result is $\frac{\partial cap_H}{\partial U_H}=sigmoid(U_H)(1-sigmoid(U_H))$. Meanwhile, $\frac{\partial U_H}{\partial W_{Z\rightarrow H}}$ and $\frac{\partial U_P}{\partial Y_H}$ also depend on the specific choice of the weighting operation $\otimes_{Z\rightarrow H}$.

Based on formulae (4)-(5), a universal backpropagation algorithm can be designed theoretically for capsule networks, with one iteration detailed in \textbf{Algorithm 1}. In practice, this algorithm should be changed to one of many variants with training data [20].

\begin{table*}
  %\caption{\textbf{Algorithm 1}. One iteration of the universal backpropagation algorithm.}
  %\label{sample-table}
  \centering
  \begin{tabular}{lll}
    \multicolumn{3}{c}{\textbf{Algorithm 1}: One iteration of the universal backpropagation algorithm.} \\
    \toprule
    1) Select a learning rate $\eta > 0$, \\
    2) $\forall H\in \mathcal{M}\cup \mathcal{O}$, $\forall Z \in IN_H$, initialize $W_{Z\rightarrow H}$ and $B_H$,  \\
    3) $\forall H\in \mathcal{O}$, compute $\delta_H = \frac{\partial Loss(Y_H,T_H)}{\partial Y_H}\cdot \frac{\partial cap_H}{\partial U_H}$, \\
    4) $\forall H \in \mathcal{M}$, compute $\delta_H = \sum_{P\in OUT_H}{\delta_P \cdot \frac{\partial U_P}{\partial Y_H}\cdot \frac{\partial cap_H}{\partial U_H}}$, \\
    5) Compute $\Delta W_{Z\rightarrow H}=\delta_H\cdot \frac{\partial U_H}{\partial W_{Z\rightarrow H}}$ and $\Delta B_H = \delta_H$, \\
    6) Update $W_{Z\rightarrow H} \leftarrow W_{Z\rightarrow H}- \eta \cdot \Delta W_{Z\rightarrow H}$,$B_H \leftarrow B_H- \eta \cdot \Delta B_H$.\\
    \bottomrule
  \end{tabular}
\end{table*}

\section{Conclusions}

Based on the formalization of neural networks, we have developed capsule networks to establish a unified framework for deep learning. This capsule framework could not only simplify the description of existing DNNs, but also provide a theoretical basis of graphical designing and programming for new deep learning models. As future work, we will try to define an industrial standard and implement a graphic platform for the advancement of deep learning with capsule networks, and even with a similar extension to recurrent neural networks.

\section*{References}

\medskip

\small

[1] Krizhevsky, A., Sutskever, I.\ \& Hinton, G.E.\ (2012) Imagenet classification with deep convolutional neural networks. In F.\ Pereira, C.J.C.\ Burges, L.\ Bottou and K.Q.\ Weinberger (eds.), {\it Advances in neural information processing systems 25}, pp.\ 1097--1105. Cambridge, MA: MIT Press.

[2] Amodei, D., Ananthanarayanan, S.\ \& Anubhai, R.\ et al.\ (2016) Deep speech 2: End-to-end speech recognition in English and Mandarin.  {\it International Conference on Machine Learning}, pp.\ 173--182.

[3] Wu, Y., Schuster, M.\ \& Chen, Z.\ et al.\ (2016) Google's Neural Machine Translation System: Bridging the Gap between Human and Machine Translation. {\it arXiv preprint arXiv:1609.08144}.

[4] Rumellhart, D.E.\ (1986) Learning internal representations by error propagation. {\it Parallel distributed processing: Explorations in the microstructure of cognition}  {\bf 1}:319-362.

[5] Schmidhuber, J.\ (2014) Deep learning in neural networks: An overview. {\it Neural Network} {\bf 61}:85-117.

[6] Hinton, G.E.\ \& Salakhutdinov, R.R.\ (2006) Reducing the dimensionality of data with neural networks. {\it Science} {\bf 313}(5786):504-507.

[7] Hinton, G.E., Osindero, S.\ \& Teh, Y.W.\ (2006) A fast learning algorithm for deep belief nets. {\it Neural computation} {\bf 18}(7):1527-1554.

[8] LeCun, Y., Bottou, L.\ \& Bengio Y, et al.\ (1998) Gradient-based learning applied to document recognition. {\it Proceedings of the IEEE} {\bf 86}(11):2278-2324.

[9] Simonyan, K.\ \& Zisserman, A.\ (2014) Very Deep Convolutional Networks for Large-Scale Image Recognition. {\it Computer Science}.

[10] Szegedy, C. Liu, W.\ \& Jia, Y.\ et al.\ (2015) Going deeper with convolutions. {\it IEEE Conference on Computer Vision and Pattern Recognition}.

[11] He, K. Zhang, X.\ \& Ren, S.\ et al.\ (2016) Deep residual learning for image recognition. {\it Proceedings of the IEEE conference on computer vision and pattern recognition}, pp.\ 770--778

[12] Ren, S., He, K.\ \& Girshick, R.\ et al.\ (2015) Faster r-cnn: Towards real-time object detection with region proposal networks. {\it Advances in neural information processing systems}, pp.\ 91--99.

[13] Huang, G., Liu, Z.\ \& Weinberger, K.Q.\ et al.\ (2017) Densely connected convolutional networks. {\it Proceedings of the IEEE conference on computer vision and pattern recognition}, pp.\ {\bf 1}(2):3.

[14] He, K., Gkioxari, G.\ \& Dollár, P.\ et al.\ (2017) Mask r-cnn. {\it Computer Vision (ICCV), 2017 IEEE International Conference on. IEEE}, pp.\ 2980--2988.

[15] Redmon, J., Divvala, S.\ \& Girshick, R.\ et al.\ (2016) You only look once: Unified, real-time object detection. {\it Proceedings of the IEEE conference on computer vision and pattern recognition}, pp.\ 779--788.

[16] Liu, W., Anguelov, D.\ \& Erhan, D.\ et al.\ (2016) Ssd: Single shot multibox detector. {\it European conference on computer vision. Springer, Cham}, pp.\ 21--37.

[17] Mnih, V., Kavukcuoglu, K.\ \& Silver, D.\ et al.\ (2015) Human-level control through deep reinforcement learning. {\it Nature}  {\bf 518}(7540):529.

[18] Silver, D., Schrittwieser, J.\ \& Simonyan, K.\ et al.\ (2017) Mastering the game of Go without human knowledge. {\it Nature} {\bf 550}(7676):354-359.

[19] Sabour, S., Frosst, N.\ \& Hinton, G.E.\ (2017) Dynamic routing between capsules. In I.\ Guyon, U.V.\ Luxburg, S.\ Bengio, H.\ Wallach, R.\ Fergus, S.\ Vishwanathan and R.\ Garnett (eds.), {\it Advances in Neural Information Processing Systems 30}, pp.\ 3859-3869. Cambridge, MA: MIT Press.

[20] Ruder, S.\ (2016) An overview of gradient descent optimization algorithms. {\it arXiv preprint arXiv:1609.04747}.

\end{document}